\def\year{2022}\relax
\documentclass[letterpaper]{article} % DO NOT CHANGE THIS
\usepackage{aaai22}  % DO NOT CHANGE THIS
\usepackage{times}  % DO NOT CHANGE THIS
\usepackage{helvet}  % DO NOT CHANGE THIS
\usepackage{courier}  % DO NOT CHANGE THIS
\usepackage[hyphens]{url}  % DO NOT CHANGE THIS
\usepackage{graphicx} % DO NOT CHANGE THIS
\usepackage{natbib}  % DO NOT CHANGE THIS AND DO NOT ADD ANY OPTIONS TO IT
\usepackage{caption} % DO NOT CHANGE THIS AND DO NOT ADD ANY OPTIONS TO IT
\DeclareCaptionStyle{ruled}{labelfont=normalfont,labelsep=colon,strut=off} % DO NOT CHANGE THIS
\usepackage{algorithm}
\usepackage{algorithmic}

\usepackage{subfigure}
\usepackage{amsfonts,amssymb}
\usepackage{amsmath}
\usepackage{algorithm}
\usepackage{algorithmic}
\usepackage{bbding}
\newcommand{\eg}{\textit{e}.\textit{g}.}
\usepackage{booktabs}

\usepackage{newfloat}
\usepackage{listings}
\floatstyle{ruled}
\newfloat{listing}{tb}{lst}{}
\floatname{listing}{Listing}
\usepackage{xcolor}

\title{Towards Transferable Adversarial Attacks on Vision Transformers}
\author{
    Zhipeng Wei\textsuperscript{\rm 1,2}\equalcontrib, 
    Jingjing Chen\textsuperscript{\rm 1,2}\equalcontrib, 
    Micah Goldblum\textsuperscript{\rm 3}, 
    Zuxuan Wu\textsuperscript{\rm 1,2}\thanks{Correspondence to: Zuxuan Wu, Yu-Gang Jiang.} \\
    Tom Goldstein\textsuperscript{\rm 3}, 
    Yu-Gang Jiang\textsuperscript{\rm 1,2\dag} 
}
\affiliations{
    \textsuperscript{\rm 1}Shanghai Key Lab of Intelligent Information Processing, School of Computer Science, Fudan University \\
    \textsuperscript{\rm 2}Shanghai Collaborative Innovation Center on Intelligent Visual Computing\\
    \textsuperscript{\rm 3}Department of Computer Science, University of Maryland\\
    
    zpwei21@m.fudan.edu.cn, chenjingjing@fudan.edu.cn, goldblumcello@gmail.com, \\
    zxwu@fudan.edu.cn, tomg@cs.umd.edu, ygj@fudan.edu.cn
}

\usepackage{bibentry}
\newcommand{\Enst}[1]{#1\ensuremath{_{ens3}}\kern-\scriptspace}
\newcommand{\Ensf}[1]{#1\ensuremath{_{ens4}}\kern-\scriptspace}
\newcommand{\Ens}[1]{#1\ensuremath{_{ens}}\kern-\scriptspace}

\begin{document}

\maketitle

\begin{abstract}
Vision transformers (ViTs) have demonstrated impressive performance on a series of computer vision tasks, yet they still suffer from adversarial examples.  % crafted in a similar fashion as CNNs. 
In this paper, we posit that adversarial attacks on transformers should be specially tailored for their architecture, jointly considering both patches and self-attention, in order to achieve high transferability.
More specifically, we introduce a dual attack framework, which contains a Pay No Attention (PNA) attack and a PatchOut attack, to improve the transferability of adversarial samples across different ViTs. We show that skipping the gradients of attention during backpropagation can generate adversarial examples with high transferability. In addition, adversarial perturbations generated by optimizing randomly sampled subsets of patches at each iteration achieve higher attack success rates than attacks using all patches. We evaluate the transferability of attacks on state-of-the-art ViTs, CNNs and robustly trained CNNs. The results of these experiments demonstrate that the proposed dual attack can greatly boost transferability between ViTs and from ViTs to CNNs.
In addition, the proposed method can easily be combined with existing transfer methods to boost performance.
Code is available at \url{https://github.com/zhipeng-wei/PNA-PatchOut}.
\end{abstract}

\section{Introduction}
 Vision Transformers (ViTs) \cite{dosovitskiy2020image} have attained excellent performance compared to state-of-the-art CNNs on standard image recognition tasks. However, ViTs are still vulnerable to security threats via adversarial examples \cite{goodfellow2014explaining, Wang2021DualAttention, chen2021attacking, wei2020heuristic}, which are nearly indistinguishable from natural images while containing perturbations that result in incorrect predictions. It is known that a model of unknown structure can be attacked using adversarial images crafted with a different ``surrogate'' model \cite{liu2016delving}.  This cross-model transferability property of adversarial examples makes it feasible to attack a ``black-box'' model without knowing its architecture or other properties.

Cross-model transferability is a well-studied phenomenon for CNNs~\cite{xie2019improving, dong2018boosting, wei2021crossmodal}. High-performance transfer attacks typically employ data augmentation and advanced gradient calculations to prevent perturbations from over-fitting to the model, as this would diminish their success when transferring to black-box models. In contrast, relatively little is known about the transferability properties of attacks crafted on ViTs, and extending existing approaches that work well on CNNs to transformers is non-trivial due to significant structural differences. More specifically, ViTs receive a sequence of flattened patches from images as inputs and use a series of multi-headed self-attention (MSA) layers to learn relationships between patches. The use of standard attacks without considering these unique architectural features will result in suboptimal results, and give the user an inaccurate sense of the adversarial vulnerability of the transformer model class.

\begin{figure}[t]
\centering
    \subfigure[Self-attention Block]{
        \centering
        \includegraphics[width=0.5\columnwidth]{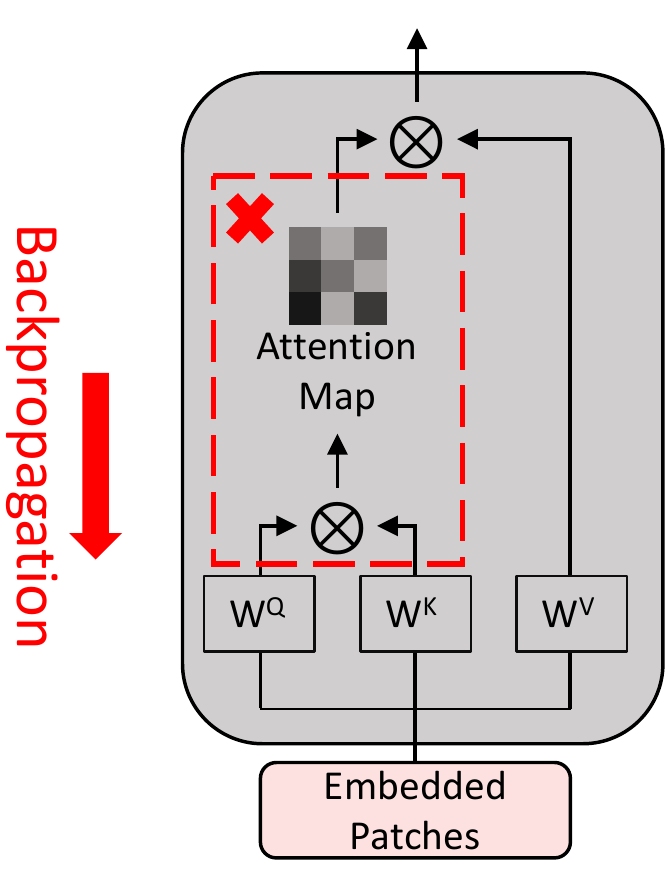}
        \label{sa_a}
    }
    \subfigure[Bypass Attention]{
        \centering
        \includegraphics[width=0.4\columnwidth]{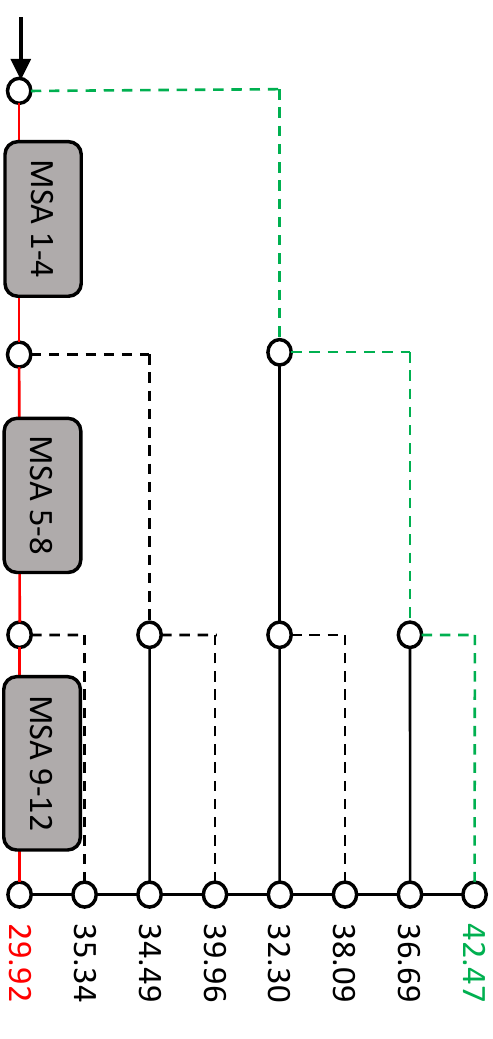}
        \label{sa_b}
    }
\caption{
(a): An illustration of the self-attention block. The attention map calculated by Q and K is dropped in gradient backpropagation (red dashed box). 
(b): Adversarial examples are generated by using gradients that bypass attention (no gradients pass through the dotted red box in Figure \ref{sa_a}). We split the 12 attention blocks of ViT-B/16 into three chunks: MSA 1-4, 5-8 and 9-12. We consider the 8 different propagation paths that include/exclude every combination of these three blocks. The rightmost path skips all attention units and achieves the best black-box attack success rate (labeled in green) while the red path through all attention blocks achieves the worst black-box attack success rate (labeled in red). The attacks are crafted by BIM on 2,000 images under perturbations with $L_{\infty}$-norm bounded above by $\epsilon=16$. The number of iterations is $10$.}
\label{ToySA}
\end{figure}

In light of these structural differences, we aim to generate highly transferable adversarial examples using white-box ViTs as proxy models to attack different black-box ViTs, normally trained CNNs, and robustly trained CNNs. In particular, we introduce a dual attack method tailored for the architecture of ViTs---attacking the attention mechanism and patches simultaneously, both of which are the core components of popular transformer architectures. In particular, we use a Pay No Attention (PNA) attack and a PatchOut attack to manipulate the attention mechanism and image features in parallel.
The PNA attack improves adversarial transferability by treating the attention weights computed on the forward pass as constants. In other words, it does not propagate through the branch of the computation graph that produces the attention weights, as illustrated in Figure \ref{sa_a}. This prevents patches from strongly interacting, as measured by the Shapley interaction index \cite{Shapley1988AVF}, which is known to help boost adversarial transferability \cite{wang2020unified}.
In addition, the PatchOut attack randomly samples a subset of patches to receive updates on each iteration of the attack crafting process. This is akin to using dropout \cite{Hinton2012ImprovingNN} on perturbation patches, and helps to combat over-fitting.  It also shares similarities with random feature selection in Random Forests \cite{Breiman2004RandomF}, Dropout and Diversity Input (DI) \cite{xie2019improving}. Besides, we integrate the $L_2$ norm \cite{chen2021going} into our dual attack to prefer a large distance from the benign sample.

\begin{figure}[h]
\centering
    \includegraphics[width=0.9\columnwidth]{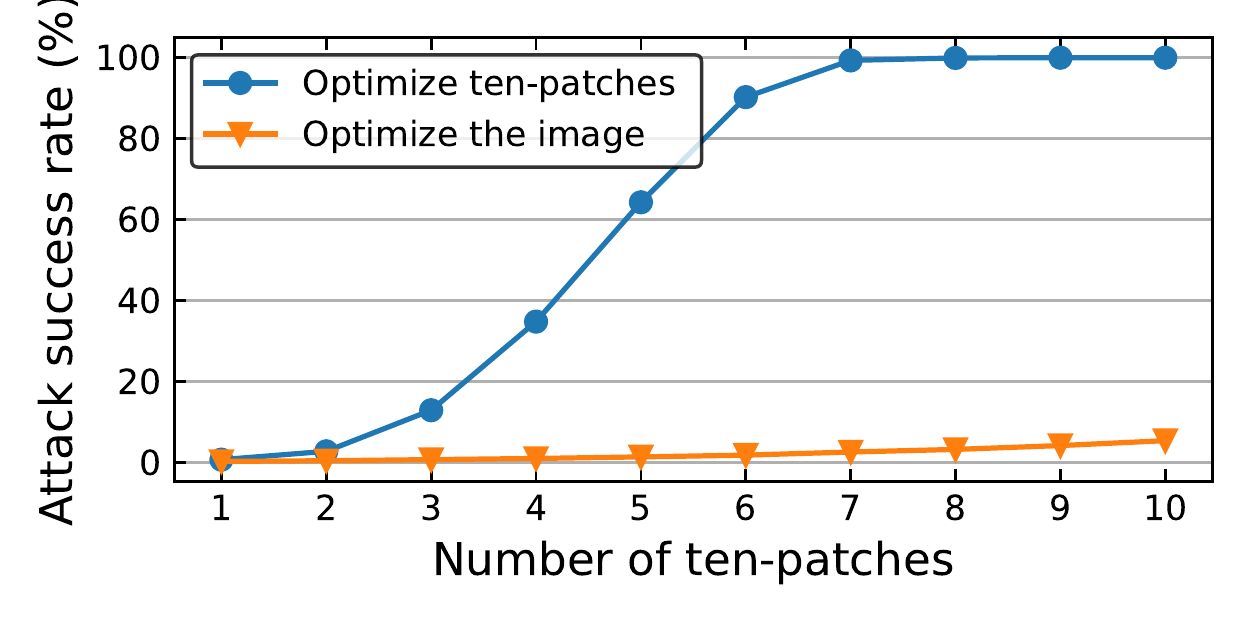}
\caption{
Attack success rate against black-box models with varying number of perturbation updates.  We consider using either ten-patch updates or whole-image updates.
The blue line generates adversarial perturbations by accumulating multiple ten-patch perturbations, which are generated by optimizing a single ten-patch at each stage. The orange line represents perturbations generated using whole-image updates. We average the attack success rate over numerous black-box models: DeiT-B, LeViT-256, CaiT-S-24, PiT-B, ConViT-B, TNT-S and Visformer-S.}
\label{ToyPP}
\end{figure}

%%%%%figure%%%%%
To validate the effectiveness of our approach, and to illustrate how the gradients of attention weights impair adversarial transferability,  we conduct two toy experiments using the BIM \cite{kurakin2016adversarial} attack on the white-box model ViT-B/16 \cite{dosovitskiy2020image} using the ImageNet validation dataset. 
Figure \ref{sa_b} shows the results of PNA. We observe that the attack success rate (ASR) decreases as more attention gradients are used during backpropagation. Bypassing all gradients of attention (the green path) improves ASR from 29.92\% to 42.47\%.
Figure \ref{ToyPP} shows the results of PatchOut, where we randomly select ten patches as one input pattern.  We call such a sparse perturbation a ``ten-patch.'' We see that stacking multiple ten-patches achieves a higher ASR than using perturbations produced by optimizing on the whole image at once. This observation demonstrates that stacking perturbations from diverse input patterns can help alleviate the over-fitting problem.

We further conduct extensive experiments on the ImageNet dataset \cite{russakovsky2015imagenet}. We  demonstrate that the proposed dual attack method significantly improves ASR against black-box ViTs, normally trained CNNs, and robustly trained CNNs. 
Compared to MI \cite{dong2018boosting} and SGM \cite{wu2020skip}, the dual attack improves ASR by 15.86\%, 27.68\%, 23.52\% on average for ViTs, normally trained CNNs and robustly trained CNNs respectively based on the adversarial transferability.
Our results demonstrate the feasibility of using white-box ViTs to attack other black-box models, and they further demonstrate the vulnerability of seemingly robustly trained CNNs.
We briefly summarize our primary contributions as follows:
\begin{itemize}
    \item We study the attention mechanism in ViTs and propose the Pay No Attention (PNA) attack to craft adversarial examples without backpropping through attention. PNA is applicable to any gradient-based attack method and any attention-based neural network.
    
    \item We study how stacking perturbations using random subsets of patches can improve attack transferability, and we propose the PatchOut Attack to generate adversarial examples using different patches as input at each iteration.
    
    \item We conduct comprehensive transfer attack experiments using 4 different white-box ViTs against 8 black-box ViTs, 4 CNNs, and 3 robustly trained CNNs, showing that the proposed dual attack can greatly improve adversarial transferability and indicating that it can be combined with existing methods to further improve performance.
\end{itemize}

\section{Related Work}
Adversarial attacks \cite{yan2020sparse} are often studied under white- and black-box threat models. The white-box setting allows the adversary fully access to victim models, while the black-box setting only permits access to the output of a victim. Hence, the black-box setting is substantially more challenging. Transfer-based attacks generate adversarial examples on white-box models with the intent that the attacked samples will also be effective against black-box models.
In this section, we review existing work on transfer-based attacks as well as on ViTs. 

\subsection{Transfer-based attacks on CNNs}
Fast Gradient Sign Method (FGSM) \cite{goodfellow2014explaining} and Basic Iterative Method (BIM) \cite{kurakin2016adversarial1} are fundamental white-box attack methods which are commonly used for transfer-based attacks. FGSM performs a one-step update in the direction of the sign of the gradient to maximizes loss. BIM applies FGSM several times iteratively with a small step size. Due to the fine-grained updates, BIM often generates more powerful adversarial examples than FGSM for white-box attacks, but BIM simultaneously achieves lower transferability than FGSM because of the over-fitting problem \cite{kurakin2016adversarial}. To overcome this problem, several efforts have been made to improve adversarial transferability. These efforts can be roughly categorized into two groups: data augmentation and advanced gradient calculation. Data augmentation ensures that the attack's effectiveness is invariant to input transformations. 
For example, the Diversity Input (DI) attack \cite{xie2019improving} applies random resizing and padding to the inputs with a fixed probability at each iteration. The Translation-Invariant (TI) attack \cite{dong2019evading} optimizes a perturbation over an ensemble of translated images by convolving the gradient with a linear or Guassian kernel, motivated by the near translation-invariance of CNNs. Scale-Invariant Method (SIM) \cite{lin2019nesterov} optimizes a perturbation over an ensemble of scaled images at each iteration.

Advanced gradient calculations stabilize the update direction or add new terms to the loss function in order to better optimize adversarial perturbations or to reduce over-fitting. For example, Momentum Iterative (MI) attack \cite{dong2018boosting} stabilizes update directions and escapes from poor local maxima by integrating a momentum term. Nesterov Accelerated Gradient \cite{lin2019nesterov} can also be viewed as another momentum term for improving transferability. Transferable Adversarial Perturbations (TAP) \cite{zhou2018transferable} maximize the distance between natural images and their adversarial examples in intermediate feature maps and also uses regularization to smooth the resulting perturbations. The Attention-Guided Transfer Attack (ATA) \cite{wu2020boosting} also maximizes distance in the attention maps obtained by Grad-CAM \cite{selvaraju2017grad} for critical feature destruction. Skip Gradient Method (SGM) \cite{wu2020skip} utilizes a decay factor to reduce gradients from the residual modules and encourages the attack to focus on more transferable low-level information. The Interaction-Reduced (IR) attack \cite{wang2020unified} discovers the negative correlation between adversarial transferability and the interactions inside perturbations. Based on the findings of IR, Wang et al. propose to decrease interactions inside perturbations, which can be implicitly represented by Shapley values. Nevertheless, the above approaches are designed for CNNs, and the performance degrades significantly when the attacks are directly applied to ViTs due to key architectural differences. 

\subsection{Vision Transformers}
Inspired by the great success of transformers on Natural Language Processing, the vision transformer (ViT) was first introduced by \cite{dosovitskiy2020image}.  This model receives raw image patches as input and is pre-trained with a large image dataset. Subsequently, many works have proposed to further improve the accuracy and efficiency of ViTs. In order to overcome the necessity of pre-training ViTs on massive datasets, DeiT \cite{touvron2021training} introduces a teacher-student strategy specific to transformers which utilizes a new distillation token to learn knowledge from CNNs. T2T-ViT \cite{yuan2021tokens} introduces a T2T module to model the local structure of the image and adopts a deep-narrow structure for the backbone of Transformers. TNT \cite{han2021transformer} utilizes an outer transformer block and an inner transformer block to learn relationships between patches and within patches separately. CaiT \cite{touvron2021going} builds deeper transformers and inserts class tokens only in later layers. CvT \cite{wu2021cvt} introduces convolutions into ViTs with the benefits of CNNs, which inserts the class token only in the last layer. Other ViTs such as LeViT \cite{graham2021levit}, PiT \cite{heo2021rethinking}, ConViT \cite{d2021convit}, Visformer \cite{chen2021visformer}, M2TR \cite{wang2021m2tr} and Deepvit \cite{zhou2021deepvit} further improve ViTs from different angles as well.
Apart from the above works, there are also some other papers \cite{tang2021robustart,paul2021vision,Shi2021DecisionbasedBA} that discuss the robustness of ViTs.

\subsection{Transfer-based attacks on ViTs}
Compared to transfer-based attacks on CNNs, less effort has been made in investigating the transferability of adversarial examples across different ViTs. One related work \cite{naseer2021improving} proposes the Self-Ensemble (SE) method, which boosts adversarial transferability by optimizing perturbations on an ensemble of models. This method utilizes the class token at each layer with a shared classification head to build an ensemble of models. As the perturbations are optimized over an ensemble of models, the generated example remains adversarial with respect to multiple models, and this property may encourage generalization to other black-box models. Additionally, \cite{naseer2021improving} also introduces a Token Refinement (TR) module to fine-tune the class tokens for further enhancing transferability. Despite exhibiting promising performance, the broad applicability of this method is limited since many ViT models do not have enough class tokens to build an ensemble.
Furthermore, TR needs to access the ImageNet training set during a fine-tuning process, which is time-consuming.  In contrast, our method is generalizable across different ViTs models and is easy to implement.

\section{Methodology}
\subsection{Preliminary}
Consider an image sample $x \in \mathcal{X} \subset \mathbb{R}^{H \times W \times C}$ and its ground-truth label $y \in \mathcal{Y} = \{1,...,K\}$, where $H$, $W$, $C$ denote height, width, and the number of channels respectively, and $K$ represents the number of classes.
We reshape the image $x$ into a sequence of flattened patches $x_{p} = \{x_p^1, ..., x_p^i, ..., x_p^N\} \in \mathbb{R}^{N \times (P^2 \cdot C)}$, where $x_p^i$ denotes the $i$-th patch of $x$, $(P, P)$ is the resolution of $x_p^i$, and $N=H \cdot W/P^2$ is the number of patches. Then, $x_{p}$ is the input to a vision transformer.
We use $f(x): \mathcal{X} \to \mathcal{Y}$ to denote the prediction function for a white-box ViT surrogate model. We use $g$ to denote a black-box victim model, which can be a vanilla CNN, robustly trained CNN, and ViT.
Following \cite{dong2018boosting, xie2019improving,wu2020skip}, we focus on untargeted adversarial attacks and enforce an $L_\infty$-norm constraint on perturbations. Hence, the goal of the transfer-based attack is to add a perturbation $\delta$ to $x$ yielding an adversarial example $x_{adv}$ using information from $f$ in order to maximize the probability that $g(x_{adv}) \neq y$ subject to the constraint that $||\delta||_{\infty} \leq \epsilon$. The optimization problem for generating adversarial examples on white-box models is formulated as follows:
\begin{equation}
    \label{eq1}
        \mathop{\arg\max}_{\delta} J(f(x+\delta),y), \\
        s.t.\,\, ||\delta||_{\infty}<\epsilon,
\end{equation}
where $J(\cdot,\cdot)$ is the loss function (\emph{e.g.} cross-entropy). 

\subsection{Pay No Attention (PNA)}
In ViTs, a single MSA uses multiple self-attention structures, where each such structure, \emph{i.e.} a head, learns its own unique features. By projecting concatenated outputs from multiple heads, MSA combines information from different representation subspaces \cite{Vaswani2017AttentionIA}.
Based on the observation that the gradients of attention in each head impair the generation of highly transferable adversarial examples, we drop the attention gradients to alleviate the interaction between patches.

Figure \ref{sa_a} illustrates the proposed Pay No Attention (PNA) attack. Given an input patch embedding $Z \in \mathbb{R}^{N \times D}$, query, key, and value weights $W^Q, W^K, W^V \in \mathbb{R}^{D \times D_h}$, the attention can be formulated as follows:
\begin{equation}
    \label{eq2}
       A = softmax(ZW^Q(ZW^K)^T/\sqrt{D_h}),
\end{equation}
where $A \in \mathbb{R}^{N \times N}$denotes the attention weight. Then the output of this head can be defined as follows:
\begin{equation}
    \label{eq3}
       Z^{'} = A(ZW^V).
\end{equation}
Thus the gradient of the output $Z^{'}$ with respect to input $Z$ can be decomposed as: 
\begin{equation}
    \frac{\partial Z^{'}}{\partial Z} = (\mathbb{I} \bigotimes A) \frac{\partial (ZW^V)}{\partial Z} + ((ZW^V)^T \bigotimes \mathbb{I}) \frac{\partial A}{\partial Z}.
\end{equation}

Our proposed PNA attack ignores backpropagation through the attention branch, i.e., it sets  $\frac{\partial A}{\partial Z} = 0$.  This is equivalent to treating the attention weights as a constant, or using a ``stop gradient'' on the attention weights.  This results in the approximation
\begin{equation}
    \frac{\partial Z^{'}}{\partial Z} \approx  (\mathbb{I} \bigotimes A) \frac{\partial (ZW^V)}{\partial Z} = (\mathbb{I} \bigotimes A) ((W^V)^T \bigotimes \mathbb{I}),
\end{equation}
where $\bigotimes$ denotes the Kronecker product. PNA forces the perturbation to exploit the network only by using feature representations, and not by exploited highly model-specific properties of attention. This results in  adversarial examples with high transferability. Skipping attention also allows gradients to focus on each patch individually,  rather than relying on complex interactions between patches. IR \cite{wang2020unified} showed a negative correlation between adversarial transferability and these multi-patch interactions. We believe this is another reason why PNA improves transferability.

\begin{algorithm}[tb]
\caption{The dual attack on ViTs}
\label{alg}
\textbf{Input}: The loss function $J$ of Equation \ref{new_loss}, a white-box model $f$, a clean image $x$ with its ground-truth class $y$.\\
\textbf{Parameter}: The perturbation budget $\epsilon$, iteration number $I$, used patch number $T$.\\
\textbf{Output}: The adversarial example.
\begin{algorithmic}[1] %[1] enables line numbers
\STATE $\delta_{0} \gets \mathbf{0}$
\STATE $\alpha \gets \frac{\epsilon}{I}$
\FOR{$i = 0$ to $I-1$}
\STATE $x_s \gets PatchOut(x_p, T)$
\STATE $M \gets \text{Equation \ref{mask}}$
\STATE $g \gets PNA(\nabla_{\delta}J$ \text{with the $L_2$ norm}) 
\STATE $\delta_{i} \gets clip_{\epsilon}({\delta_{i-1} + \alpha \cdot g})$
\ENDFOR
\STATE $x_{adv} = x + \delta_I$
\STATE \textbf{return} $x_{adv}$ 
\end{algorithmic}
\end{algorithm}

\begin{table*}[ht]
    \centering
    \begin{tabular}{l c c c c c c c c}
        \toprule
         Method & ViT-B/16 & PiT-B & CaiT-S-24 & Visformer-S & DeiT-B & TNT-S & LeViT-256 & ConViT-B  \\ \midrule
         FGSM & 15.57 & 19.80 & 20.43 & 19.37 & 22.08 & 22.78 & 18.80 & 25.58\\ 
         BIM & 20.77 & 22.17 & 22.63 & 22.70 & 33.53 & 32.13 & 20.45 & 35.30 \\ 
         MI & 41.23 &	45.23 &	47.13 &	45.97 & 56.03 & 55.23 & 43.75 &	58.25	  \\ 
         DI & 32.57 & 45.13 & 43.07 & 47.77 & 48.08 & 55.18 & 43.25 &	49.35	  \\ 
         TI & 19.33 & 17.67 & 16.50 & 19.00 & 25.13 & 28.18 & 13.70 & 27.53  \\ 
         SIM & 34.97 & 32.73 & 35.17 & 31.13 & 44.13 & 46.73 & 36.43 & 45.68  \\ 
         SGM & 38.87 & 41.60 & 52.30 & 48.80 & 60.53 & 64.33 & 51.13 & 60.68  \\ 
         IR & 21.33 & 22.70 & 24.00 & 23.43 & 34.00 & 33.43 & 21.30 & 36.38  \\ 
         TAP & 25.27 & 24.73 & 33.40 & 32.20 & 43.20 & 39.78 & 30.03 & 42.20  \\ 
         ATA & \ \ 3.47 & \ \ 1.13 & \ \ 0.97 & \ \ 2.67 & \ \ 3.68 & \ \ 3.37 & \ \ 2.02 & \ \ 3.72 \\ 
         SE & 29.05 & 21.25 & 31.40 & 24.90 & 45.23 & 37.87 & 21.73 & 46.03  \\ 
         Ours & \textbf{46.10} & \textbf{52.40} & \textbf{59.87} & \textbf{58.60} & \textbf{63.85} & \textbf{67.25} & \textbf{57.62} & \textbf{63.70}  \\ \bottomrule
    \end{tabular}
    \caption{
    MASR (\%) against ViTs with various attack methods. We use ViT-B/16, PiT-B, CaiT-S-24 and Visformer-S as white-box models respectively to generate adversarial examples. Each model is evaluated on adversarial examples generated by white-box ViTs, and the ASRs are averaged over surrogate models to obtain MASR. The best results are in \textbf{bold}.}
    \label{tab_vits}
\end{table*}

\subsection{PatchOut Attack}
To motivate our work, we consider the findings of \cite{xie2019improving}, which suggests that diverse input patterns can improve the transferability of adversarial examples by alleviating the over-fitting phenomenon.
Therefore, we introduce the PatchOut Attack, which randomly attacks a subset of patches at each iteration to alleviate over-fitting.

We use $T$ to control the number of used patches at each iteration, and $x_{s} = \{ x_s^1, ..., x_s^t, ..., x_s^T \}$ to denote the selected patches. Thus, the attack mask $M \in \{ 0,1 \}^{H \times W \times C }$ can be formulated as:
\begin{equation}
    \label{mask}
        M_p^i = 
        \begin{cases}
    1,              & \text{if } x_p^i \text{ in }  x_s\\
    0,              & \text{otherwise}
        \end{cases}
\end{equation}
where $M_p^i \in \{ 0,1 \}^{P \times P \times C}$ is the region corresponding to $x_p^i$ in the image. Therefore, PatchOut replaces Equation \ref{eq1} with: 
\begin{equation}
    \label{new_loss}
    \mathop{\arg\max}_{\delta} J(f(x+ M \odot \delta),y) + \lambda||\delta||_2, \\
        s.t.\,\, ||\delta||_{\infty}<\epsilon,
\end{equation}
where $\odot$ denotes element-wise multiplication. The added second term encourages perturbations to have a large $L_2$ norm, preferring a large distance from $x$. $\lambda$ controls the balance between the loss function and the regularization term.

We finally summarize the proposed dual attack to craft adversarial examples in Algorithm \ref{alg}, where the function $PatchOut(\cdot, \cdot)$ randomly select $T$ patches from $x_p$ to generate $x_s$, the function $PNA(\cdot)$ bypasses the gradients of attention, and $clip_{\epsilon}(\cdot)$ restricts each entry of generated perturbations to be within $[-\epsilon, \epsilon]$.

\section{Experiments}
\subsection{Experimental Settings}
\textbf{Dataset.} 
Following the setting from \citet{dong2018boosting, dong2019evading, xie2019improving, lin2019nesterov}, we randomly sample one image, which is correctly classified by all models, from each class from the ImageNet 2012 validation dataset \cite{russakovsky2015imagenet}, to conduct our experiments.

\textbf{Models.} 
We evaluate the performance and transferability of the proposed method under two different settings: 1) The surrogate and victim models are both ViTs; 2) The surrogate is a ViT, and the victim model is a CNN. In these experiments, we evaluate performance on multiple ViT variants as well as multiple CNNs architectures, including both normally trained models and robustly trained models. For ViTs, we conduct experiments on ViT-B/16 \cite{dosovitskiy2020image}, DeiT-B \cite{touvron2021training}, TNT-S \cite{han2021transformer}, LeViT-256 \cite{graham2021levit}, PiT-B \cite{heo2021rethinking}, CaiT-S-24 \cite{touvron2021going}, ConViT-B \cite{d2021convit}, and Visformer-S \cite{chen2021visformer}. While for CNNs, the experiments are conducted on Inception v3 (Inc-v3) \cite{szegedy2016rethinking}, Inception v4 (Inc-v4), Inception ResNet v2 (IncRes-v2) \cite{szegedy2017inception}, and ResNet v2-152 (Res-v2) \cite{he2016identity}. We chose these models to conduct our experiments since they are publicly available in the timm library \cite{rw2019timm}. We randomly select ViT-B/16, PiT-B, CaiT-S-24, Visformer-S as the white-box models to generate adversarial examples. When selecting one model as the white-box surrogate, we use the remaining models as black-box victims when evaluating performance.

\textbf{Evaluation.}
We use Attack Success Rate (ASR) to quantify performance on black-box models. Here, ASR denotes the proportion of generated adversarial examples from the surrogate model that are successfully misclassified by the black-box victim model. A higher success rate means better adversarial transferability. We also use Mean ASR (MASR) to denote the mean value of ASR for one black-box victim, averages across different white-box surrogate models.
Following \citet{dong2018boosting, dong2019evading, xie2019improving}, we set the norm constraint $\epsilon=16$ and $J$ as the cross-entropy loss function. For the iterative attack, we set $I=10$ and thus the step size $\alpha=1.6$.
We resize all images to $224 \times 224$ to conduct experiments.
For the inputs of ViTs, we set the patch size $P=16$, thus the number of the patches is $N=196$.

\begin{table*}[]
    \centering
    \begin{tabular}{l c c c c c c c}
        \toprule
         Method & Inc-v3 & Inc-v4 & IncRes-v2 & Res-v2 & \Enst{Inc-v3} & \Ensf{Inc-v3} & \Ens{IncRes-v2}  \\ \midrule
         FGSM & 20.78 & 18.80 & 16.38 & 19.05 & 14.62 & 13.98 & 10.38	 \\ 
         BIM & 17.88 & 14.77 & 12.40 & 11.82 & \ \ 8.02 & \ \ 6.20 & \ \ 4.45 \\ 
         MI & 39.65 & 37.43 & 32.17 & 33.28 & 24.80 & 22.10 & 17.68	\\ 
         DI & 32.78 & 31.75 & 26.40 & 25.00 & 17.73 & 15.22 & 11.12	\\ 
         TI & 23.27 & 23.60 & 15.28 & 20.88 & 18.80 & 20.80 & 13.92  \\ 
         SIM & 30.55 & 27.63 & 24.17 & 24.48 & 19.45 & 17.67 & 13.40 \\ 
         SGM & 38.42 & 34.00 & 27.25 & 27.25 & 18.70 & 16.53 & 11.68 \\ 
         IR & 17.65 & 15.83 & 12.08 & 12.48 & \ \ 8.05 & \ \ 6.50 & \ \ 4.78  \\ 
         TAP & 29.58 & 26.10 & 20.67 & 19.23 & 12.58 & 10.62 & \ \ 6.85 \\ 
         ATA & \ \ 3.25 & \ \ 2.53 & \ \ 2.03 & \ \ 2.07 & \ \ 1.20 & \ \ 0.85 & \ \ 0.92 \\ 
         SE & 18.40 & 16.47 & 12.33 & 12.63 & \ \ 9.13 & \ \ 6.73 & \ \ 5.10  \\ 
         Ours & \textbf{47.95} & \textbf{45.12} & \textbf{38.45} & \textbf{38.93} & \textbf{26.20} & \textbf{22.85} & \textbf{18.10}  \\ \bottomrule
    \end{tabular}
    \caption{MASR (\%) against CNNs with various attack methods. We use ViT-B/16, PiT-B, CaiT-S-24 and Visformer-S as white-box models to generate adversarial examples. Each CNN is evaluated on adversarial examples generated by white-box ViTs, and the ASRs are averaged over surrogate models to obtain MASR. The best results are in \textbf{bold}.}
    \label{tab_cnns}
\end{table*}

\subsection{Performance Comparison}
We compare our method with several baseline attacks, including MI \cite{dong2018boosting}, DI \cite{xie2019improving}, TI \cite{dong2019evading}, SIM \cite{lin2019nesterov}, SGM \cite{wu2020skip}, IR \cite{wang2020unified}, TAP \cite{zhou2018transferable}, ATA \cite{wu2020boosting}, and SE \cite{naseer2021improving}, which are integrated into BIM \cite{kurakin2016adversarial}. Note that we do not compare with TR \cite{naseer2021improving} because it requires fine-turning on ImageNet, and thus would yield an unfair comparison in terms of both data access and compute cost. 
For each baseline method, we follow their original settings in our experiments. 
We set our patch number $T=130$, and we set the balancing parameter $\lambda = 0.1$ in PatchOut according to the experimental results. 
For performance comparison, we report the MASR obtained from white-box models including ViT-B/16, PiT-B, CaiT-S-24 and Visformer-S.

\textbf{Performance on ViTs.} We first evaluate the adversarial transferability of our method across different ViTs. Table \ref{tab_vits} summarizes the results on different black-box ViT models. From the results, we have the following observations.
First, TI and ATA, which improve the adversarial transferability significantly on CNNs, exhibit poor performances on ViTs.
As TI is proposed based on the property of translation invariance in CNNs, it is not applicable to ViTs since the architecture of ViTs leads to much less image-specific inductive bias than CNNs (\eg, translation invariance) \cite{yuan2021tokens}. Furthermore, while the key idea of ATA is to corrupt the features of discriminative regions shared among different CNNs, it is also not applicable to ViTs for the reason that the discriminative region varies significantly across different ViTs. 
Second, compared to other baseline methods designed for CNNs, SGM performs much better on ViTs. This is because the key idea of SGM is to boost transferability by encouraging the model to focus more on low-level information, which is also applicable to ViTs since SGM is not designed based on the unique characteristics of CNNs. 
Third, SE, which is proposed for improving the transferability of ViTs, does not work well across different structures of ViTs. This is because the performance of SE is closely related to the number of class tokens for building the ensemble models. As there are few or even no class tokens in some ViTs (e.g, PiT-B, Visformer-S), SE works poorly for such situations.  
Lastly, our dual attack consistently outperforms the baseline attacking methods and achieves a 58.67 \% average MASR by averaging MASR across different black-box models. Our method, which considers the core components of ViTs, achieves the best results in ViTs with different structures. These experiments validate the effectiveness of the proposed dual attack.

\textbf{Performance on CNNs.}
We further evaluate the performance of transferring the adversarial examples generated on white-box ViTs to attack black-box CNNs. The results are summarized in Table \ref{tab_cnns}. We have the following observations. First, the attack success rates of all methods decrease significantly, which indicates that adversarial samples generated on ViTs are less transferable on CNNs than ViTs, presumably due to structural differences between ViTs and CNNs. Second, MI improves the transferability of the generated adversarial examples by utilizing the momentum term to stabilize update directions. 
Third, compared to other baseline methods, our proposed dual attack significantly improves MASR despite the different structures between ViTs and CNNs. The dual attack has a 42.61\% average MASR on normally trained CNNs and a 22.38\% average MASR on robustly trained CNNs. These results indicate the feasibility of using ViTs to attack robustly trained CNNs.

\textbf{Combining with Existing Methods.}
The proposed dual attack tailored for ViTs can be easily combined with existing methods to further improve performance. We demonstrate this by combining our method with MI and SGM. The reason we choose MI and SGM is that these two methods perform much better than other baseline methods. 
Using these combined methods, we conduct experiments with ViT-B/16 as the white-box model. Specifically, SGM + Ours uses fewer gradients than MultiLayer Perceptron (MLP) modules according to a decay factor that is set as $0.5$.
Table \ref{tab_combine} shows the results of the integrated methods.
We observe that the proposed dual attack significantly improves the transferability of MI and SGM across ViTs, normally trained CNNs, and robustly trained CNNs. In particular, the dual attack increases the average MASR of MI and SGM by 28.54 \% and 102.49 \%, respectively, when attacking robustly trained CNNs.
The results demonstrate that the dual attack can easily be combined with existing methods to further boost performance.

\begin{figure}[t]
    \centering
    \includegraphics[width=0.8\columnwidth]{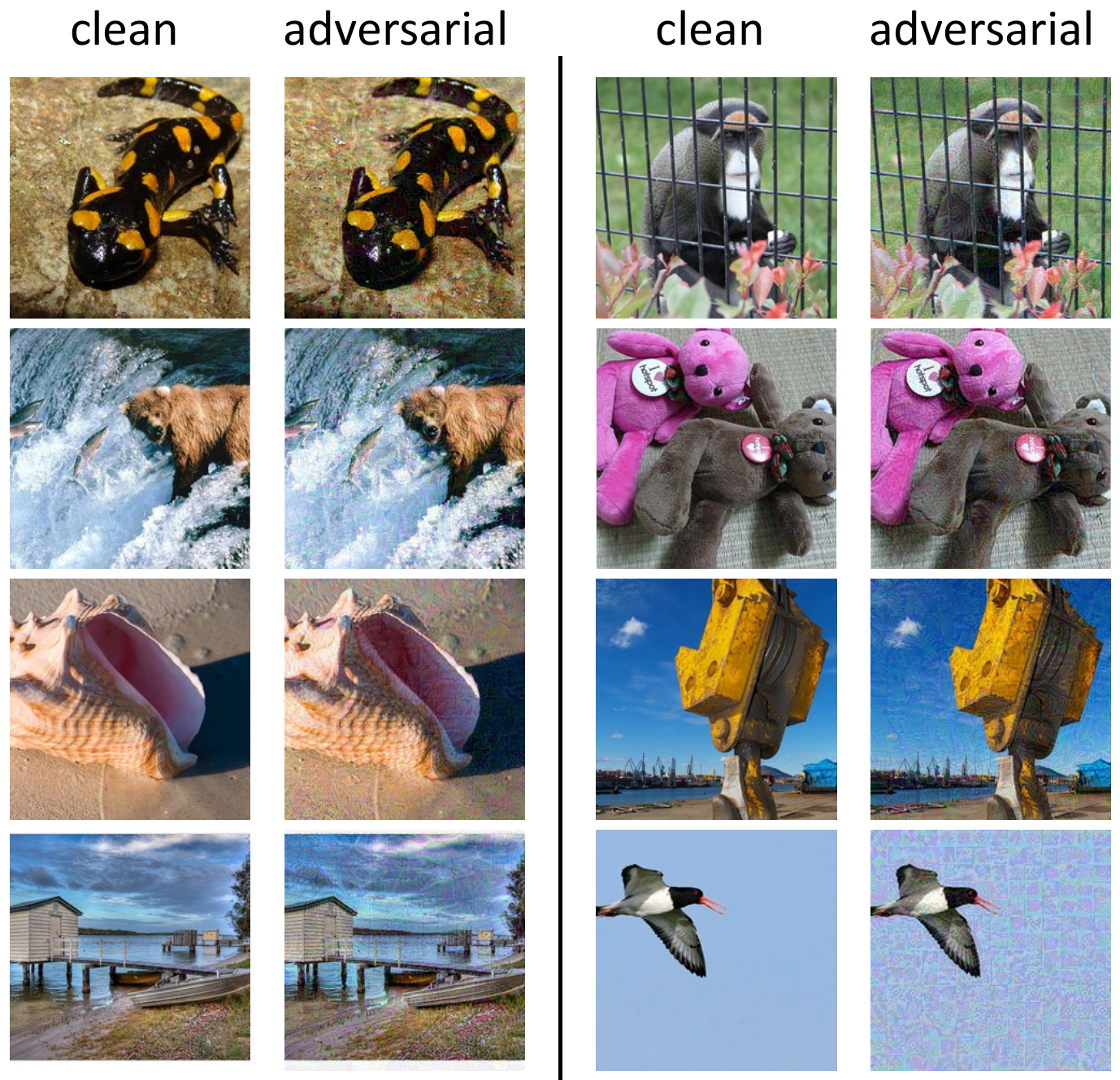}
    \caption{Visualization of randomly picked clean images and their corresponding adversarial images, crafted on the ViT-B/16 model using our dual attack.}
    \label{fig_show}
\end{figure}

\textbf{Visualization of Adversarial Examples.}
Figure \ref{fig_show} depicts 8 randomly selected clean images and their corresponding adversarial examples. Here, we observe that the adversarial perturbations are hardly perceptible.

\begin{table}[]
    \centering
    \begin{tabular}{l c c c}
        \toprule
         Method & ViTs & Normal CNNs & Robust CNNs \\ \midrule
         MI & 49.10 & 35.63 & 21.53 \\ 
         MI + Ours & 59.70 & 48.79 & 30.13\\ 
         SGM & 52.28 &  31.73 & 15.64\\ 
         SGM + Ours & 63.48 & 52.31 & 31.37\\ \bottomrule
    \end{tabular}
    \caption{Average ASR (\%) of models under attacks that combine the proposed method and existing algorithms. The experiment is conducted by using ViT-B/16 as the white-box model. Average ASR is obtained by averaging ASRs of different types of black-box models. "Normal CNNs" and "Robust CNNs" denote normally trained CNNs and robustly trained CNNs respectively.}
    \label{tab_combine}
\end{table}

\subsection{Ablation Studies}
To validate the effect of each component proposed in our method, we evaluate the performance under various combinations of the components: PatchOut, the regularization term $L_2$ in Equation \ref{new_loss}, and Pay No Attention (PNA). We also conduct experiments to analyze the effect of hyper-parameters, including the used patch number $T$ and the balance parameter $\lambda$. For the ablation study, we use ViT-B/16 as the white-box model and generate adversarial examples on 2,000 randomly sampled images to attack other black-box models.

\textbf{Effect of Components.}
Table \ref{tab_abl} shows the results of different component combinations. It is observed that using any of the components leads to performance improvements on both ViTs and CNNs. 
Among these components, PNA improves adversarial transferability most significantly on both ViTs and CNNs, demonstrating that gradients from ViTs, except for those from attention modules, are most vulnerable.
The best performance is achieved when combining all components together, which suggests that the three components contribute to the improvement of adversarial transferability in complementary ways.

\begin{table}[]
    \centering
    \begin{tabular}{c c c c c c}
        \toprule
         PatchOut & $L_2$ & PNA & ViTs & Normal & Robust  \\ \midrule
         - & - & - & 26.97 & 10.63 & \ \ 5.50\\ 
         \checkmark & - & - & 34.92 & 12.55 & \ \ 7.07\\ 
         - & \checkmark & - & 37.79 & 16.55 & 10.00\\ 
         - & - & \checkmark & 42.47 & 18.75 & 10.97\\ 
         \checkmark & \checkmark & - & 48.82 & 24.55 & 14.67\\ 
         - & \checkmark & \checkmark & 49.69 & 22.36 & 13.54\\
         \checkmark & \checkmark & \checkmark & 59.15 & 36.23 & 23.87\\ \bottomrule
    \end{tabular}
    \caption{Average ASR (\%) for our proposed method with different component combinations. PatchOut indicates the operation of sampling a subset of patches, $L_2$ denotes the regularization term in Equation \ref{new_loss}, PNA is the operation that bypasses the gradients of attention. `\checkmark' indicates that the component is used while `-' indicates that it is not used.  ``Normal" and ``Robust" denote normally trained CNNs and robustly trained CNNs respectively.}
    \label{tab_abl}
\end{table}

\begin{figure}[t]
\centering
    \subfigure[The used patch number $T$]{
        \centering
        \includegraphics[width=0.85\columnwidth]{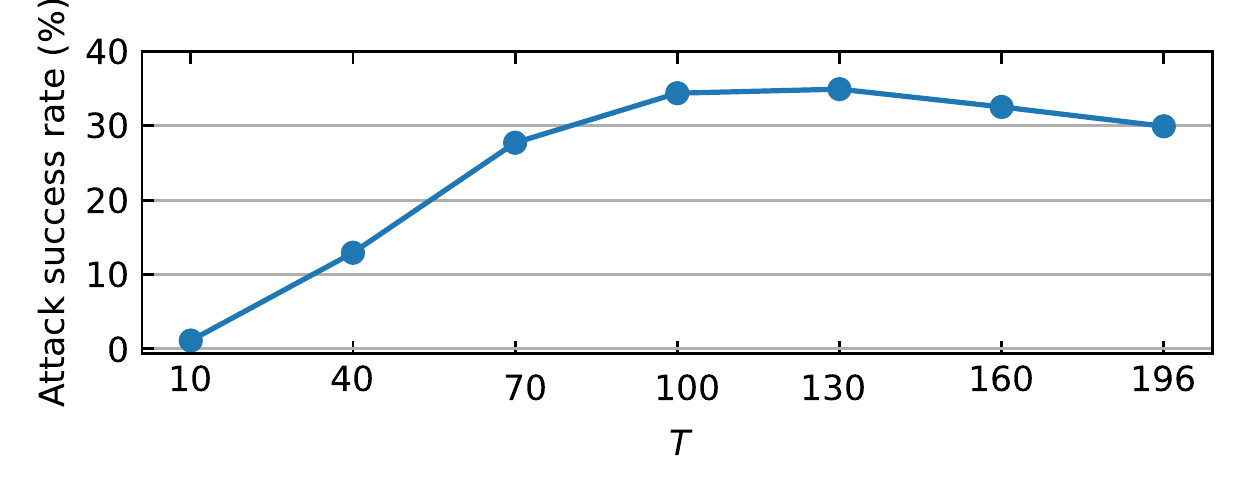}
        \label{com_ucf}
    }
    \subfigure[$\lambda$]{
        \centering
        \includegraphics[width=0.85\columnwidth]{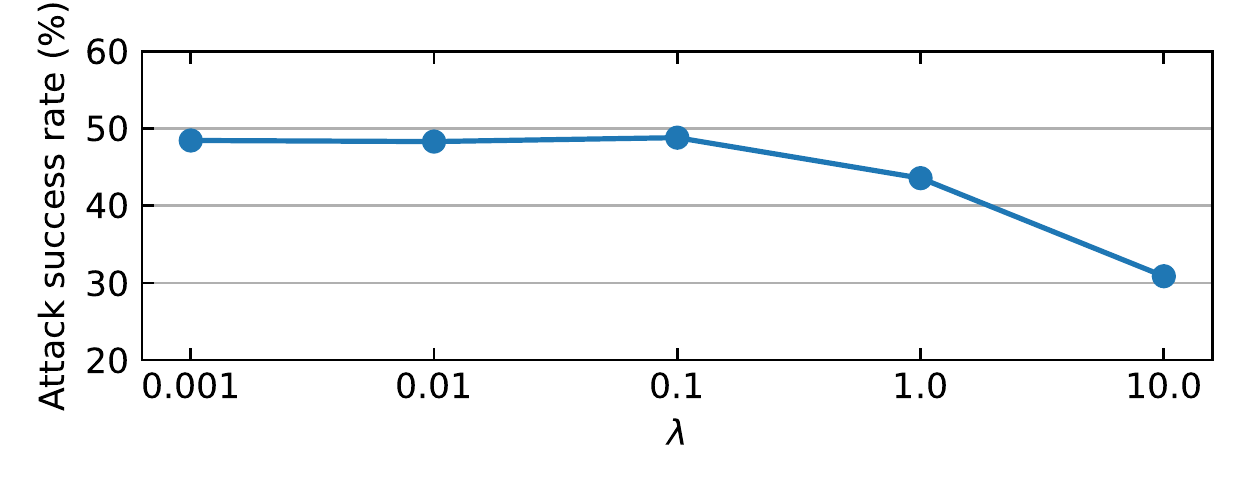}
        \label{com_kinetics}
    }
\caption{Average ASR (\%) of ViTs under attacks using various values of (a) patch number $T$ and (b) $\lambda$ in Equation \ref{new_loss}.}
\label{Paras}
\end{figure}

\textbf{Effect of Parameters.}
We investigate the effect of parameters $T$ and $\lambda$ and summarize the results in Figure \ref{Paras}. 
The used patch number $T$ determines how many patches are used in each attack iteration. When all patches are included ($T=N$), PatchOut degenerates into BIM. When $T$ is small, PatchOut is not able to generate strong adversarial examples in a limited number of iterations ($I=10$). Thus, it is vital to study optimal settings of $T$.
We experiment by tuning $T$ without $L_2$. Figure \ref{com_ucf} shows the results, with $T \in [10, 40, 70, 100, 130, 160, 196]$. When $T=130$, PatchOut achieves optimal results. With this value of $T$, each patch can be attacked multiple times in the context of different patch subsets.
For the balance parameter $\lambda$, we set $\lambda \in [0.001, 0.01, 0.1, 1, 10]$ to conduct the experiments with $T=130$. Figure \ref{com_kinetics} illustrates the effect of various $\lambda$. When $\lambda = 0.1$, PatchOut achieves the best result by balancing the contribution of each term in Equation \ref{new_loss}.

\section{Conclusion}
In this paper, we identify several properties of ViTs that can be leveraged to produce more transferable adversarial examples. Specifically, we find that ignoring the gradients of attention units and only perturbing a subset of the patches at each iteration prevents overfitting and creates diverse input patterns, thus increasing transferability. To validate these assumptions, we conduct two toy experiments. After verifying our intuitions, we propose the dual attack for ViTs consisting of the Pay No Attention (PNA) attack and the PatchOut attack to craft adversarial examples with high transferability.
We conduct a series of experiments with 8 ViTs, 4 normally trained CNNs, and 3 robustly trained CNNs to show that the proposed method can greatly improve adversarial transferability. Combining our attack with other existing methods, our techniques consistently enhance adversarial transferability. 
The results of this study show that cross-model transferability happens even across very wide architectural gaps between models. The transferability observed in this work further suggests that the feature extractors and implicit biases of transformer architectures and CNNs are not as different as one might suspect, and we think adversarial examples might serve as the basis for future studies of the similarities and differences between the two.

\section{Acknowledgment}
The authors would like to thank the anonymous referees for their valuable comments and helpful suggestions. This work was supported in part by NSFC project (\#62032006), Science and Technology Commission of Shanghai Municipality Project (20511101000), and in part by Shanghai Pujiang Program (20PJ1401900).

\bibliography{refer.bib}

\begin{thebibliography}{44}
\providecommand{\natexlab}[1]{#1}

\bibitem[{Breiman(2004)}]{Breiman2004RandomF}
Breiman, L. 2004.
\newblock Random Forests.
\newblock \emph{Machine Learning}, 45: 5--32.

\bibitem[{Chen et~al.(2021{\natexlab{a}})Chen, Wei, Chen, Wu, and
  Jiang}]{chen2021attacking}
Chen, K.; Wei, Z.; Chen, J.; Wu, Z.; and Jiang, Y.-G. 2021{\natexlab{a}}.
\newblock Attacking Video Recognition Models with Bullet-Screen Comments.
\newblock \emph{arXiv preprint arXiv:2110.15629}.

\bibitem[{Chen et~al.(2021{\natexlab{b}})Chen, Tao, Ye, and
  Huang}]{chen2021going}
Chen, S.; Tao, Q.; Ye, Z.; and Huang, X. 2021{\natexlab{b}}.
\newblock Going Far Boosts Attack Transferability, but Do Not Do It.
\newblock \emph{arXiv preprint arXiv:2102.10343}.

\bibitem[{Chen et~al.(2021{\natexlab{c}})Chen, Xie, Niu, Liu, Wei, and
  Tian}]{chen2021visformer}
Chen, Z.; Xie, L.; Niu, J.; Liu, X.; Wei, L.; and Tian, Q. 2021{\natexlab{c}}.
\newblock Visformer: The vision-friendly transformer.
\newblock \emph{arXiv preprint arXiv:2104.12533}.

\bibitem[{d'Ascoli et~al.(2021)d'Ascoli, Touvron, Leavitt, Morcos, Biroli, and
  Sagun}]{d2021convit}
d'Ascoli, S.; Touvron, H.; Leavitt, M.; Morcos, A.; Biroli, G.; and Sagun, L.
  2021.
\newblock Convit: Improving vision transformers with soft convolutional
  inductive biases.
\newblock \emph{arXiv preprint arXiv:2103.10697}.

\bibitem[{Dong et~al.(2018)Dong, Liao, Pang, Su, Zhu, Hu, and
  Li}]{dong2018boosting}
Dong, Y.; Liao, F.; Pang, T.; Su, H.; Zhu, J.; Hu, X.; and Li, J. 2018.
\newblock Boosting adversarial attacks with momentum.
\newblock In \emph{Proceedings of the IEEE conference on computer vision and
  pattern recognition}, 9185--9193.

\bibitem[{Dong et~al.(2019)Dong, Pang, Su, and Zhu}]{dong2019evading}
Dong, Y.; Pang, T.; Su, H.; and Zhu, J. 2019.
\newblock Evading defenses to transferable adversarial examples by
  translation-invariant attacks.
\newblock In \emph{Proceedings of the IEEE/CVF Conference on Computer Vision
  and Pattern Recognition}, 4312--4321.

\bibitem[{Dosovitskiy et~al.(2020)Dosovitskiy, Beyer, Kolesnikov, Weissenborn,
  Zhai, Unterthiner, Dehghani, Minderer, Heigold, Gelly
  et~al.}]{dosovitskiy2020image}
Dosovitskiy, A.; Beyer, L.; Kolesnikov, A.; Weissenborn, D.; Zhai, X.;
  Unterthiner, T.; Dehghani, M.; Minderer, M.; Heigold, G.; Gelly, S.; et~al.
  2020.
\newblock An image is worth 16x16 words: Transformers for image recognition at
  scale.
\newblock \emph{arXiv preprint arXiv:2010.11929}.

\bibitem[{Goodfellow, Shlens, and Szegedy(2014)}]{goodfellow2014explaining}
Goodfellow, I.~J.; Shlens, J.; and Szegedy, C. 2014.
\newblock Explaining and harnessing adversarial examples.
\newblock \emph{arXiv preprint arXiv:1412.6572}.

\bibitem[{Graham et~al.(2021)Graham, El-Nouby, Touvron, Stock, Joulin,
  J{\'e}gou, and Douze}]{graham2021levit}
Graham, B.; El-Nouby, A.; Touvron, H.; Stock, P.; Joulin, A.; J{\'e}gou, H.;
  and Douze, M. 2021.
\newblock LeViT: a Vision Transformer in ConvNet's Clothing for Faster
  Inference.
\newblock \emph{arXiv preprint arXiv:2104.01136}.

\bibitem[{Han et~al.(2021)Han, Xiao, Wu, Guo, Xu, and
  Wang}]{han2021transformer}
Han, K.; Xiao, A.; Wu, E.; Guo, J.; Xu, C.; and Wang, Y. 2021.
\newblock Transformer in transformer.
\newblock \emph{arXiv preprint arXiv:2103.00112}.

\bibitem[{He et~al.(2016)He, Zhang, Ren, and Sun}]{he2016identity}
He, K.; Zhang, X.; Ren, S.; and Sun, J. 2016.
\newblock Identity mappings in deep residual networks.
\newblock In \emph{European conference on computer vision}, 630--645. Springer.

\bibitem[{Heo et~al.(2021)Heo, Yun, Han, Chun, Choe, and
  Oh}]{heo2021rethinking}
Heo, B.; Yun, S.; Han, D.; Chun, S.; Choe, J.; and Oh, S.~J. 2021.
\newblock Rethinking spatial dimensions of vision transformers.
\newblock \emph{arXiv preprint arXiv:2103.16302}.

\bibitem[{Hinton et~al.(2012)Hinton, Srivastava, Krizhevsky, Sutskever, and
  Salakhutdinov}]{Hinton2012ImprovingNN}
Hinton, G.~E.; Srivastava, N.; Krizhevsky, A.; Sutskever, I.; and
  Salakhutdinov, R. 2012.
\newblock Improving neural networks by preventing co-adaptation of feature
  detectors.
\newblock \emph{ArXiv}, abs/1207.0580.

\bibitem[{Kurakin, Goodfellow, and Bengio(2016)}]{kurakin2016adversarial}
Kurakin, A.; Goodfellow, I.; and Bengio, S. 2016.
\newblock Adversarial machine learning at scale.
\newblock \emph{arXiv preprint arXiv:1611.01236}.

\bibitem[{Kurakin et~al.(2016)Kurakin, Goodfellow, Bengio
  et~al.}]{kurakin2016adversarial1}
Kurakin, A.; Goodfellow, I.; Bengio, S.; et~al. 2016.
\newblock Adversarial examples in the physical world.

\bibitem[{Lin et~al.(2019)Lin, Song, He, Wang, and Hopcroft}]{lin2019nesterov}
Lin, J.; Song, C.; He, K.; Wang, L.; and Hopcroft, J.~E. 2019.
\newblock Nesterov accelerated gradient and scale invariance for adversarial
  attacks.
\newblock \emph{arXiv preprint arXiv:1908.06281}.

\bibitem[{Liu et~al.(2016)Liu, Chen, Liu, and Song}]{liu2016delving}
Liu, Y.; Chen, X.; Liu, C.; and Song, D. 2016.
\newblock Delving into transferable adversarial examples and black-box attacks.
\newblock \emph{arXiv preprint arXiv:1611.02770}.

\bibitem[{Naseer et~al.(2021)Naseer, Ranasinghe, Khan, Khan, and
  Porikli}]{naseer2021improving}
Naseer, M.; Ranasinghe, K.; Khan, S.; Khan, F.~S.; and Porikli, F. 2021.
\newblock On Improving Adversarial Transferability of Vision Transformers.
\newblock \emph{arXiv preprint arXiv:2106.04169}.

\bibitem[{Paul and Chen(2021)}]{paul2021vision}
Paul, S.; and Chen, P.-Y. 2021.
\newblock Vision transformers are robust learners.
\newblock \emph{arXiv preprint arXiv:2105.07581}.

\bibitem[{Russakovsky et~al.(2015)Russakovsky, Deng, Su, Krause, Satheesh, Ma,
  Huang, Karpathy, Khosla, Bernstein et~al.}]{russakovsky2015imagenet}
Russakovsky, O.; Deng, J.; Su, H.; Krause, J.; Satheesh, S.; Ma, S.; Huang, Z.;
  Karpathy, A.; Khosla, A.; Bernstein, M.; et~al. 2015.
\newblock Imagenet large scale visual recognition challenge.
\newblock \emph{International journal of computer vision}, 115(3): 211--252.

\bibitem[{Selvaraju et~al.(2017)Selvaraju, Cogswell, Das, Vedantam, Parikh, and
  Batra}]{selvaraju2017grad}
Selvaraju, R.~R.; Cogswell, M.; Das, A.; Vedantam, R.; Parikh, D.; and Batra,
  D. 2017.
\newblock Grad-cam: Visual explanations from deep networks via gradient-based
  localization.
\newblock In \emph{Proceedings of the IEEE international conference on computer
  vision}, 618--626.

\bibitem[{Shapley(1988)}]{Shapley1988AVF}
Shapley, L. 1988.
\newblock A Value for n-person Games.

\bibitem[{Shi and Han(2021)}]{Shi2021DecisionbasedBA}
Shi, Y.; and Han, Y. 2021.
\newblock Decision-based Black-box Attack Against Vision Transformers via
  Patch-wise Adversarial Removal.

\bibitem[{Szegedy et~al.(2017)Szegedy, Ioffe, Vanhoucke, and
  Alemi}]{szegedy2017inception}
Szegedy, C.; Ioffe, S.; Vanhoucke, V.; and Alemi, A.~A. 2017.
\newblock Inception-v4, inception-resnet and the impact of residual connections
  on learning.
\newblock In \emph{Thirty-first AAAI conference on artificial intelligence}.

\bibitem[{Szegedy et~al.(2016)Szegedy, Vanhoucke, Ioffe, Shlens, and
  Wojna}]{szegedy2016rethinking}
Szegedy, C.; Vanhoucke, V.; Ioffe, S.; Shlens, J.; and Wojna, Z. 2016.
\newblock Rethinking the inception architecture for computer vision.
\newblock In \emph{Proceedings of the IEEE conference on computer vision and
  pattern recognition}, 2818--2826.

\bibitem[{Tang et~al.(2021)Tang, Gong, Wang, Liu, Wang, Chen, Yu, Liu, Song,
  Yuille et~al.}]{tang2021robustart}
Tang, S.; Gong, R.; Wang, Y.; Liu, A.; Wang, J.; Chen, X.; Yu, F.; Liu, X.;
  Song, D.; Yuille, A.; et~al. 2021.
\newblock Robustart: Benchmarking robustness on architecture design and
  training techniques.
\newblock \emph{arXiv preprint arXiv:2109.05211}.

\bibitem[{Touvron et~al.(2021{\natexlab{a}})Touvron, Cord, Douze, Massa,
  Sablayrolles, and J{\'e}gou}]{touvron2021training}
Touvron, H.; Cord, M.; Douze, M.; Massa, F.; Sablayrolles, A.; and J{\'e}gou,
  H. 2021{\natexlab{a}}.
\newblock Training data-efficient image transformers \& distillation through
  attention.
\newblock In \emph{International Conference on Machine Learning}, 10347--10357.
  PMLR.

\bibitem[{Touvron et~al.(2021{\natexlab{b}})Touvron, Cord, Sablayrolles,
  Synnaeve, and J{\'e}gou}]{touvron2021going}
Touvron, H.; Cord, M.; Sablayrolles, A.; Synnaeve, G.; and J{\'e}gou, H.
  2021{\natexlab{b}}.
\newblock Going deeper with image transformers.
\newblock \emph{arXiv preprint arXiv:2103.17239}.

\bibitem[{Vaswani et~al.(2017)Vaswani, Shazeer, Parmar, Uszkoreit, Jones,
  Gomez, Kaiser, and Polosukhin}]{Vaswani2017AttentionIA}
Vaswani, A.; Shazeer, N.~M.; Parmar, N.; Uszkoreit, J.; Jones, L.; Gomez,
  A.~N.; Kaiser, L.; and Polosukhin, I. 2017.
\newblock Attention is All you Need.
\newblock \emph{ArXiv}, abs/1706.03762.

\bibitem[{Wang et~al.(2021{\natexlab{a}})Wang, Liu, Yin, Liu, Tang, and
  Liu}]{Wang2021DualAttention}
Wang, J.; Liu, A.; Yin, Z.; Liu, S.; Tang, S.; and Liu, X. 2021{\natexlab{a}}.
\newblock Dual Attention Suppression Attack: Generate Adversarial Camouflage in
  Physical World.
\newblock In \emph{IEEE Conference on Computer Vision and Pattern Recognition}.

\bibitem[{Wang et~al.(2021{\natexlab{b}})Wang, Wu, Chen, and
  Jiang}]{wang2021m2tr}
Wang, J.; Wu, Z.; Chen, J.; and Jiang, Y.-G. 2021{\natexlab{b}}.
\newblock M2TR: Multi-modal Multi-scale Transformers for Deepfake Detection.
\newblock \emph{arXiv preprint arXiv:2104.09770}.

\bibitem[{Wang et~al.(2020)Wang, Ren, Lin, Zhu, Wang, and
  Zhang}]{wang2020unified}
Wang, X.; Ren, J.; Lin, S.; Zhu, X.; Wang, Y.; and Zhang, Q. 2020.
\newblock A unified approach to interpreting and boosting adversarial
  transferability.
\newblock \emph{arXiv preprint arXiv:2010.04055}.

\bibitem[{Wei et~al.(2020)Wei, Chen, Wei, Jiang, Chua, Zhou, and
  Jiang}]{wei2020heuristic}
Wei, Z.; Chen, J.; Wei, X.; Jiang, L.; Chua, T.-S.; Zhou, F.; and Jiang, Y.-G.
  2020.
\newblock Heuristic black-box adversarial attacks on video recognition models.
\newblock In \emph{Proceedings of the AAAI Conference on Artificial
  Intelligence}, volume~34, 12338--12345.

\bibitem[{Wei et~al.(2021)Wei, Chen, Wu, and Jiang}]{wei2021crossmodal}
Wei, Z.; Chen, J.; Wu, Z.; and Jiang, Y.-G. 2021.
\newblock Cross-Modal Transferable Adversarial Attacks from Images to Videos.
\newblock arXiv:2112.05379.

\bibitem[{Wightman(2019)}]{rw2019timm}
Wightman, R. 2019.
\newblock PyTorch Image Models.
\newblock \url{https://github.com/rwightman/pytorch-image-models}.

\bibitem[{Wu et~al.(2020{\natexlab{a}})Wu, Wang, Xia, Bailey, and
  Ma}]{wu2020skip}
Wu, D.; Wang, Y.; Xia, S.-T.; Bailey, J.; and Ma, X. 2020{\natexlab{a}}.
\newblock Skip connections matter: On the transferability of adversarial
  examples generated with resnets.
\newblock \emph{arXiv preprint arXiv:2002.05990}.

\bibitem[{Wu et~al.(2021)Wu, Xiao, Codella, Liu, Dai, Yuan, and
  Zhang}]{wu2021cvt}
Wu, H.; Xiao, B.; Codella, N.; Liu, M.; Dai, X.; Yuan, L.; and Zhang, L. 2021.
\newblock Cvt: Introducing convolutions to vision transformers.
\newblock \emph{arXiv preprint arXiv:2103.15808}.

\bibitem[{Wu et~al.(2020{\natexlab{b}})Wu, Su, Chen, Zhao, King, Lyu, and
  Tai}]{wu2020boosting}
Wu, W.; Su, Y.; Chen, X.; Zhao, S.; King, I.; Lyu, M.~R.; and Tai, Y.-W.
  2020{\natexlab{b}}.
\newblock Boosting the transferability of adversarial samples via attention.
\newblock In \emph{Proceedings of the IEEE/CVF Conference on Computer Vision
  and Pattern Recognition}, 1161--1170.

\bibitem[{Xie et~al.(2019)Xie, Zhang, Zhou, Bai, Wang, Ren, and
  Yuille}]{xie2019improving}
Xie, C.; Zhang, Z.; Zhou, Y.; Bai, S.; Wang, J.; Ren, Z.; and Yuille, A.~L.
  2019.
\newblock Improving transferability of adversarial examples with input
  diversity.
\newblock In \emph{Proceedings of the IEEE/CVF Conference on Computer Vision
  and Pattern Recognition}, 2730--2739.

\bibitem[{Yan, Wei, and Li(2020)}]{yan2020sparse}
Yan, H.; Wei, X.; and Li, B. 2020.
\newblock Sparse black-box video attack with reinforcement learning.
\newblock \emph{arXiv preprint arXiv:2001.03754}.

\bibitem[{Yuan et~al.(2021)Yuan, Chen, Wang, Yu, Shi, Jiang, Tay, Feng, and
  Yan}]{yuan2021tokens}
Yuan, L.; Chen, Y.; Wang, T.; Yu, W.; Shi, Y.; Jiang, Z.; Tay, F.~E.; Feng, J.;
  and Yan, S. 2021.
\newblock Tokens-to-token vit: Training vision transformers from scratch on
  imagenet.
\newblock \emph{arXiv preprint arXiv:2101.11986}.

\bibitem[{Zhou et~al.(2021)Zhou, Kang, Jin, Yang, Lian, Jiang, Hou, and
  Feng}]{zhou2021deepvit}
Zhou, D.; Kang, B.; Jin, X.; Yang, L.; Lian, X.; Jiang, Z.; Hou, Q.; and Feng,
  J. 2021.
\newblock Deepvit: Towards deeper vision transformer.
\newblock \emph{arXiv preprint arXiv:2103.11886}.

\bibitem[{Zhou et~al.(2018)Zhou, Hou, Chen, Tang, Huang, Gan, and
  Yang}]{zhou2018transferable}
Zhou, W.; Hou, X.; Chen, Y.; Tang, M.; Huang, X.; Gan, X.; and Yang, Y. 2018.
\newblock Transferable adversarial perturbations.
\newblock In \emph{Proceedings of the European Conference on Computer Vision
  (ECCV)}, 452--467.

\end{thebibliography}

\end{document}